\documentclass[10pt]{article}

\usepackage{acl2017}
\usepackage{pslatex}

\newcommand\citeA{\newcite}

\usepackage{amsmath}
\usepackage{caption}
\usepackage{xcolor}
\usepackage{graphicx}
\usepackage{multirow}
\usepackage{pgfplots}
\usepackage{subcaption}
\usepackage{tabularx}
\usepackage{hhline}

\usepackage{times}
\usepackage{url}

\newcommand\ie{\emph{i.e.}}
\newcommand\eg{\emph{e.g.}}

\newcommand{\ignore}[1]{}

\newcommand{\Table}[1]{Table~\ref{#1}}

\newcommand{\Section}[1]{Section~\ref{#1}}

\newcommand{\Figure}[1]{Figure~\ref{#1}}

\newcommand\todo[1]{\textcolor{orange}{#1}}

\newcommand\sxs[1]{\textcolor{red}{#1}}
\newcommand\fm[1]{\textcolor{black}{#1}}

\title{Predicting and Explaining Human Semantic Search in a Cognitive Model}

\author{Filip Miscevic \\
   Cognitive Science Program, \\
  Complex Networks \& Systems \\
  Indiana University Bloomington \\
  {\tt fmiscevi@iu.edu} \\\And
  Aida Nematzadeh \\
  Department of Psychology \\
  University of California  \\
  Berkeley \\
  {\tt nematzadeh@berkeley.edu} \\ \And 
  Suzanne Stevenson\\
   Department of Computer Science\\
  University of  Toronto \\
  {\tt suzanne@cs.toronto.edu} \\}

\aclfinalcopy
\def\aclpaperid{12}

\begin{document}

\maketitle

\begin{abstract}
Recent work has attempted to characterize the structure of semantic memory and the search algorithms which, together, best approximate human patterns of search revealed in a semantic fluency task.
There are a number of models that seek to capture semantic search processes over networks, but they vary in the cognitive plausibility of their implementation. Existing work has also neglected to consider the constraints that the incremental process of language acquisition must place on the structure of semantic memory. Here we present a model that incrementally updates a semantic network, with limited computational steps, and replicates many patterns found in human semantic fluency using a simple random walk.
We also perform thorough analyses showing that a combination of both structural and semantic features are correlated with human performance patterns.

\end{abstract}

\section{Human Semantic Processing}
The study of human semantic memory---word meanings, their relations, and their storage---is challenging due to the complexity of factors involved. Finding (1) the right representation for word meanings and their relations, (2) the mechanism responsible for learning the representation, (3) the appropriate search algorithm to efficiently retrieve information from semantic memory, and (4) the suitable empirical data to evaluate the proposed representations and algorithms is a difficult task.
Previous research has extensively explored each of these areas \cite[\eg,][]{collins.loftus.1975, steyvers.tenenbaum.2005, griffiths.etal.2007}.

Psychologists frequently use a task known as semantic fluency (or verbal fluency) to examine human semantic representation and processing \cite{troyer.etal.1997,ardila.etal.2006}. Participants are asked to produced as many words as they can from a given category (\eg, animal) in a fixed amount of time (\eg, three minutes). The resulting data---which words people recall and in what order---can shed light on how people represent words and meanings and their relations, and how they search such semantic information.
For example, \citeA{hills.etal.2012} find that participants tend to reply in semantically-related bursts of words---$\eg$, they recall words from the pet subcategory of animals (\textit{dog}, \textit{cat}) then switch to a different subcategory, such as African animals (\textit{lion}, \textit{zebra}), etc.---indicating that people tend to follow a strategy of \textit{exploiting} a semantically-related patch of words, then \textit{exploring} to find a new patch, \fm{much like animals foraging for patches of food  in their environment}.

Recent work has investigated the properties of semantic representations and processing algorithms that can account for this type of behavior in the semantic fluency task.
Different researchers have found that a match to human behavior can be achieved in either of two ways: (a) using a simple (vector-based) semantic representation in combination with an informed, two-stage algorithm to exploit and explore the space \cite{hills.etal.2012}; or (b) creating a richer representation---structured as a semantic network---and using a simple random walk to access it \cite{abbott.etal.2015,nematzadeh.etal.2016}.
These findings suggest that the choice of representation and search algorithm are interdependent, such that the same empirical data can be replicated through different combinations of representation and algorithm that make different trade-offs on the locus of complexity \cite{abbott.etal.2015}.

However, if both combinations account for the human data considered thus far, the question of which model more plausibly captures what occurs in a search in human semantic memory remains open. As \citeA{abbott.etal.2015} suggest, further experiments, such as those performed by \citeA{hills.etal.2015}, can help elucidate the differences between these approaches to modelling human semantic memory. In particular, if there are key aspects of human semantic search that can be explained by one model and not the other, then this goes towards disconfirming the latter. One of the goals of the current paper is to show that a random walk over a semantic network does, in fact, reproduce additional empirical patterns of the human semantic fluency task.
%

In addition to these experimental approaches, other findings and theoretical considerations may come to bear on resolving the question of which model most aptly reflects human semantic search.

For example, people appear to have a structured semantic memory that encodes many kinds of relational knowledge \cite{MillerFellbaum91}. 
In this way, complexity costs are incurred during learning (while creating the structured representation) rather than every time the representations are accessed. As such, accessing the knowledge later becomes a more efficient process. Hence, it may be reasonable to suggest that a simple search algorithm operating over a structured semantic network is a preferable model.
%

Another open issue is precisely what kind of semantic representations realistically capture word relations, especially semantic similarity, which typically form the basic structure of a semantic network \cite[\eg,][]{MillerFellbaum91}.
Work modeling human semantic fluency behavior using a simple random walk over a semantic network has drawn on several different kinds of semantic word representations.
\citeA{abbott.etal.2015} constructed their semantic network using human association norms \cite{nelson.etal.1998}, so that weighted edges between words directly capture the similarities between them that are relevant to the fluency task \cite{jones.etal.2015}.
\citeA{nematzadeh.etal.2016} built two networks based on different semantic representations learned from text corpora: a simple vector-based representation model, called BEAGLE, learned from Wikipedia  \cite[previously used by \citeA{hills.etal.2012}]{jones.etal.2007}, and probability distributions learned from child-directed corpora \cite{fazly.etal.2010.csj}. 
%
Given that a random walk over semantic networks from each of these sources---human association norms, vector-space representations, and probability distributions---all model human fluency behavior, how do we choose between them? 

An important set of considerations that we explore here involves the cognitive plausibility of how a semantic representation could be learned.
While the human association norms used by \citeA{abbott.etal.2015} accurately capture human judgments of word relatedness, it is not clear how the similarity assessments captured in such norms can be learned through exposure to language.
The BEAGLE vector-space representations, on the other hand, are learned from instances of natural language. However, acquisition is a batch process over Wikipedia data, which is arguably not a good proxy for the linguistic input from which individuals acquire their semantic lexicon.
The probability distributions used by \citeA{fazly.etal.2010.csj}, on the other hand, are learned by a cognitive model from a corpus of child-directed speech. These representations thus meet important criteria for cognitive plausibility, in that they are learned from naturalistic linguistic input.

\fm{One final} crucial issue that has remained unaddressed to date is the incremental learnability of the semantic network structure itself. 
%
Children simultaneously learn word meanings as well as the relations between them \cite{jones.etal.1991}.
Thus, it is important to model the simultaneous incremental learning of both semantic word representations and their structure in a semantic network.
This has been neglected by previous work discussed so far. Even in the work where semantic representations are learned, only the word representations and not their relations are learned. Instead, the semantic network is created by exhaustively comparing all the word representations after training---a process that is too computationally demanding to be cognitively plausible. 

Our contributions in this paper are threefold:
First, we show that a semantic network created incrementally within an online word learning model---from naturalistic child language acquisition data---can yield human performance in semantic search using a simple random walk. 
Our work here confirms that a semantic network created and updated incrementally---while the model is learning words---has the appropriate structure to yield patterns observed in the semantic fluency task, despite having noisy and incomplete connections as a result of being generated from partial knowledge acquired at each time step. 
Second, as mentioned, we show that the new approach to creating the semantic network yields a structure that also mimics other aspects of human behavior in semantic fluency, going beyond earlier models in the scope of empirical data accounted for \cite{abbott.etal.2015,nematzadeh.etal.2016}. 

Finally, we go beyond previous analyses of semantic organization to determine more precisely which network properties are correlated with the observed human performance patterns.
While other work has focused on the importance of structural properties of the network in determining human behavior \cite{goni.etal.2010,steyvers.tenenbaum.2005},
we find that both structural \emph{and} semantic properties are necessary to generate patterns observed in human semantic fluency data.
\section{Incremental Network Creation}
\label{sec:model}

We use the approach of \citeA{nematzadeh.etal.2014.emnlp} to incrementally build a semantic network, which draws on the probabilistic cross-situational word learning model developed by \citeA{fazly.etal.2010.csj}.

\subsection{Incremental Word Learning Model}
\label{sec:inc-learner}

The semantic network is generated from word meanings (representations) learned by the model of \citeA{fazly.etal.2010.csj}, trained on the Manchester corpus \cite{theakston.etal.2001} of the CHILDES database \cite{macwhinney.2000}. Each input to the model consists of an \textit{utterance} from the corpus, labelled with a \textit{scene} consisting of semantic features for each word. 
For example, consider the following utterance (U) and selected features from its accompanying scene (S):
\begin{table}[h!]
\small{
\vspace{-0.05cm}
\begin{tabular}{l}
{\bf $U$:} $\{$\emph{look}, \emph{at}, \emph{the}, \emph{monkey}, \emph{eat}, \emph{a}, \emph{banana}$\}$ \\
{\bf $S$:} $\{$ \dots, \textsc{vertebrate}, \textsc{mammal}, \dots,
\textsc{fruit}, \dots $\}$
\end{tabular}
}
\vspace{-0.25cm}
\end{table}

\noindent
Just as a child must learn the referent of each word in a sentence, the learner must infer which features in the scene are associated---or \textit{aligned}---with each word.  The model captures this association as the probability of a feature $f$ given a word $w$, $P(f|w)$, which it incrementally updates from the co-occurrence of $f$ with $w$ across all observed utterance--scene pairs. The meaning of each word $w$ is then represented as the probability distribution $P(\cdot|w)$ over all semantic features, which is estimated through latent variables that model the possible alignments of words and features in an utterance--scene pair. An incremental Expectation Maximization algorithm is used to update $P(\cdot|w)$ \cite{neal.hinton.1998}. Hence, as in children, word meanings are gradually learned after many exposures to utterances and scenes.

In particular, for a single utterance--scene pair processed at time $t$, the alignment ($a$) probability of each feature ($f_i$) in the scene and word in the utterance ($w_j$) is calculated by:
\begin{align*}
    P_t(a_{ij}|f_i)&= \frac{P_{t-1}(f_i|w_j)}{\sum_{w'\in u} P_{t-1}(f_i|w') }
\end{align*}
$P_{t=0}(f_i|w_j)$ is initially randomly uniformly distributed. Once the alignment probabilities are calculated, the word meanings are updated: 
\begin{align*}
    P_t(f_i|w_j)&= \frac{\sum_{u\in U_t} P_t(a_{ij}|u,f_i)}{\sum_{f'\in M_t}\sum_{u\in U_t} P_t(a_{ij}|u,f') }
\end{align*}
Here, $U_t$ represents the set of utterances processed up to and including time $t$, and $M_t$ is the set of features observed up to and including time $t$. Note that the summations do not have to be calculated anew each time; the terms from the first $t-1$ utterances can be stored and updated with the contributions from the $t^{th}$ utterance--scene pair.

The learned representation for a word, $P(\cdot | w)$, can be treated as a vector representation of the word over all semantic features.
In the present study, we focus on animal nouns, as they are the target of the semantic fluency task in humans. The semantic features of noun meanings used are derived from WordNet hypernyms \cite[\url{http://wordnet.princeton.edu}]{fellbaum.1998}, and embed hierarchical conceptual knowledge of nouns. 

The more features (hypernyms, in this case) two animal words (\eg, \textsc{``cat",``dog"} vs. \textsc{``cat",``frog"}) have in common, the more similar their learned representations will be. The model learns not only the features associated with that particular word, however, but also features that often occur in the same context as the word. For example, in the above utterance--scene pair, the model may come to associate a non-zero probability with the feature \textsc{fruit} and the word \emph{monkey}. Hence, the learned meanings of words capture not only a conceptual hierarchy for that word but also information learned from the context of their usage.

\subsection{Incremental Learning of Semantic Networks}
\label{sec:inc-network}

Children do not just learn the meanings of words, they also learn the relations between them at the same time \cite{jones.etal.1991}. We use the approach taken by \citeA{nematzadeh.etal.2014.emnlp} to enable the model to learn word meanings and the relationships between them simultaneously, without exhaustively considering all possible relationships between the words.

Since the probability distribution $P(\cdot |w)$ for a given word $w$ is stored as a vector over all semantic features, the cosine of the angle between them can be computed as a measure of their similarity. A semantic network can thus be constructed by representing each word as a node in the network, with an edge between them if the cosine similarity between two words is greater than a threshold $\rho$.

Whenever a new utterance--scene pair $U$--$S$ is processed, the probabilities $P(\cdot | w_u)$ of all $w_u \in U$ are updated, affecting the cosine similarities between words $w_u$ and all other words.  The semantic network must be updated to reflect these changes in cosine similarities---\ie, some edges may be added, some removed, some changed in weight. However, rather than calculating the (new) cosine similarities between each $w_u$ and \textit{all} other words, the model use a limited set of calculations.
It first updates the current edges connecting $w_u$ to its neighbors. Then it selects a small set of new words $w_i$ that \textit{potentially} have a high probability of being similar to $w_u$. 
This is accomplished by incrementally forming semantic clusters over word meanings that are adjusted when a word's meaning is updated \cite{anderson.matessa.1992}.
Each newly updated word meaning $w_u$ is compared to an average (\ie, prototype) representation of each cluster to determine its probability of belonging to that cluster. Finally, $n$ words are selected from each cluster and their cosine similarity to $w_u$ updated, where $n$ is proportional to the probability of $w_u$ belonging to that cluster. The number of computations is limited as $w_u$ is only compared to the cluster prototypes and a restricted number of words from each cluster.

By limiting the number of computations at each step of learning, the model is more cognitively plausible than exhaustively updating the semantic network after each utterance.  However, it also means that the resulting semantic network will be noisy---it may have missing, superfluous or incorrectly-weighted edges.
\section{Experimental Data and Approach}

In this section, we explain the details of the semantic fluency experiment as well as the semantic representation and search algorithm used in our simulations. \fm{All of the code and data necessary to reproduce our experiments is available at \url{https://github.com/FilipMiscevic/random_walk}.}

\subsection{Evaluation: Semantic Fluency Data}

We evaluate our simulations using data from a semantic fluency experiment in which participants were tasked with naming as many animals as they can in three minutes \cite{hills.etal.2012,hills.etal.2015}. \citeA{hills.etal.2012} inferred that the recalled words (\eg, \textit{dog}, \textit{cat}, \textit{lion}, \textit{zebra}) form semantically-related categories or ``patches'',
based on their inter-item retrieval times (IRT)---the time elapsed between the naming of two sequential items that have not previously been recalled. They find that the IRT increases as search within a semantically-related category progresses.  A switch into a different semantic category occurs when the IRT exceeds the participant's average IRT across the entire trial. The IRT then decreases and the pattern begins again (see Figure \ref{fig:hills_irt}). This result shows that participants exhibit different behavior when recalling words from within a semantic category compared to switching into a new semantic category. \citeA{hills.etal.2012} argue that this pattern is a consequence of an informed two-stage search process: local cues, such as similarity to the most recent response, are used to search within patches, and global cues, such as the overall frequency of a word, are used to switch into new patches.
Here we replicate previous results that demonstrate that the IRT pattern (\Figure{fig:hills_irt}) can be predicted by a simple search given structured representations \cite{abbott.etal.2015,nematzadeh.etal.2016}. In addition, we show that this process matches other patterns observed in the semantic fluency experiment \cite{hills.etal.2015}.

\subsection{Representation: A Semantic Network}
\label{sec:semnet-rep}

We assume words and their relations are structured as a semantic network---a graph whose nodes are words, and edges reflect the similarity between the word meanings. We compare two sets of semantic networks, one set created \textit{after training} the word learner explained in \Section{sec:inc-learner}, while the other is built \textit{incrementally during the training}, as described in \Section{sec:inc-network}.
While the model learns many words, we only consider animal words, as we can evaluate those against the semantic fluency experiment of \citeA{hills.etal.2012}. We also include the word \emph{animal} itself in the semantic networks, as this is the cue word used in the experiment.

Two words $w_i$ and $w_j$ are connected in the semantic network if the cosine similarity between their feature vectors, $P(\cdot | w_i)$ and $P(\cdot | w_j)$, is above the threshold, $\rho=0.8$.
An exception is made for words connected to the word \emph{animal}: because \emph{animal} is a hypernym of the other animals, its cosine similarity will be less than the cosine between animals of the same subcategory. As such, to ensure that \emph{animal} remains connected to some words in the network, edges radiating from it are kept if the similarity is at least $\rho_{animal}=0.4$. Both models learn the representations of all 93 animal words present in the corpus; however, not all nodes are guaranteed to be connected to the rest of the network due to this thresholding.
These thresholds were determined by a grid search over the possible values of $\rho$ and $\rho_{animal}$ (\ie, $(0,1]$). The model predicts the human data over a notable range of parameter values; nonetheless, there are still more networks in that parameter space that do not predict the data. In \Section{sec:explain}, we will explore what characteristics of the networks are responsible for their successful prediction of data.
\ignore{\todo{Filip, did you vary ro-animal similar to ro? I think our not-predicting-irt networks are not really all the non-predicting-irt networks -- they are the subset that exhibit SW structure. But, I may be remembering this wrong. Just make sure my writing is correct.} \fm{Yes, $\rho_{animal}$ was varied in the same way, so this is accurate.}}

\noindent\textbf{Batch Network.} The word learner was trained on 120k utterance--scene pairs, with the meaning representation of a word, $P(\cdot | w)$, calculated as described in \Section{sec:inc-learner}. After training has concluded, a semantic network is constructed using the final learned representations. A total of 70 words is present in this network.

\noindent\textbf{Incremental Network.}  The learner is trained on 28k utterance--scene pairs.\footnote{Even with the smaller corpus (28k as opposed to 120k input pairs), the model predicts the semantic fluency data; thus, we used the smaller corpus to speed up our simulations.} After each utterance--scene pair is processed, the connections in the semantic network are updated as described in \Section{sec:inc-network}. A total of 75 words is present in this network.

Note that although the word representations of each model are learned by the same learning algorithm, they produce very different semantic networks. In the Batch Network, the edges are created only after training is completed, and is accomplished by exhaustively computing the cosine similarity between all word-pair combinations. The Incremental Network, on the other hand, uses a more cognitively plausible approximation of this process whereby edges are incrementally created by comparing only a small percentage of the word pairs.\footnote{This ends up being only $8\%$ of all $\frac{n(n-1)}{2}$ possible comparisons at each time step, where $n$ is the total number of words seen by the learner at each time step.}
This means that relations captured by the edges of the Incremental Network are noisier and incomplete.

The Incremental Network still only approximates the process of semantic acquisition in people, albeit more plausibly compared with previous work. As described above, however, we empirically set two thresholds that determine whether words are connected or not: one for the word \textit{animal} and another one for all other animal words. Future work will need to explore whether this distinction can be learned while the network is incrementally created.

\subsection{Search Algorithm: A Random Walk}
We model the search process as a random walk in which semantic information is retrieved by randomly visiting nodes in the semantic network. 
Recall that in the semantic fluency experiment, the participants were cued by the word \emph{animal} and were asked to name as many animals they can in three minutes. 
Following \citeA{abbott.etal.2015}, we simulate this experiment by performing a weighted random walk on each network, beginning with the word \emph{animal}. At each step in the random walk, a neighboring node is visited with a probability proportional to the edge weight connecting them, and the visited word is stored. 
Just as repeated words are not considered in the human recall data, we assume the output of a random walk to be the sequence of unique words encountered---\ie, each word is counted in the output only when retrieved for the first time. 
The number of steps taken before the walk terminates (including steps to already-visited nodes) is 70, which produces about the same number of words on the networks as human participants on average do (\ie, $37\pm 5$). The results we report are averages over 300 such walks.

\subsection{Analyzing Random Walks}

In the semantic fluency task, the human response patterns are reflected in changes in the inter-item retrieval time (IRT) over the list of responses.  In the empirical data, IRT is  
the time elapsed from one word until the next word is recalled, and increases and decreases are observed as people switch from one semantic patch of words to another, as noted above. 
Thus, to evaluate the random walks in our semantic networks against this IRT pattern,
we must define a measure of time in the simulated walks (since actual model speed is not an appropriate proxy).
We also must determine what constitutes a patch and a switch between two patches.

\subsubsection{Measuring Time and Semantic Distance}

We follow \citeA{abbott.etal.2015} in defining the IRT in a random walk on a semantic network as the number of steps taken (i.e., number of edges crossed) between two words. More specifically, we define IRTs for our walks as follows: for each word that has not previously been visited by the random walk, the IRT is the number of steps taken in the random walk since the last word that was seen for the first time. For example, if the model visits the sequence of nodes ``\textsc{cat,dog,cat,rat}'', the random walk output is ``\textsc{cat,dog,rat}'', and the IRT between \textsc{cat} and \textsc{dog} is 1, whereas the IRT between \textsc{dog} and \textsc{rat} is 2.

The IRT is considered a proxy for semantic distance between the words. \citeA{hills.etal.2015} also looked directly at semantic distances in the sequences generated in the human fluency task: They used vector-space representations (of the BEAGLE model) to calculate cosine similarity between consecutive words. 
%
As such, in addition to using 
IRT in assessing our walks,
we also draw on the 
cosine similarities between words.
%

\subsubsection{Identifying Patch Switches}
\begin{figure}  

\centering
  \includegraphics[trim=0 0 0 0, clip,scale=0.26]{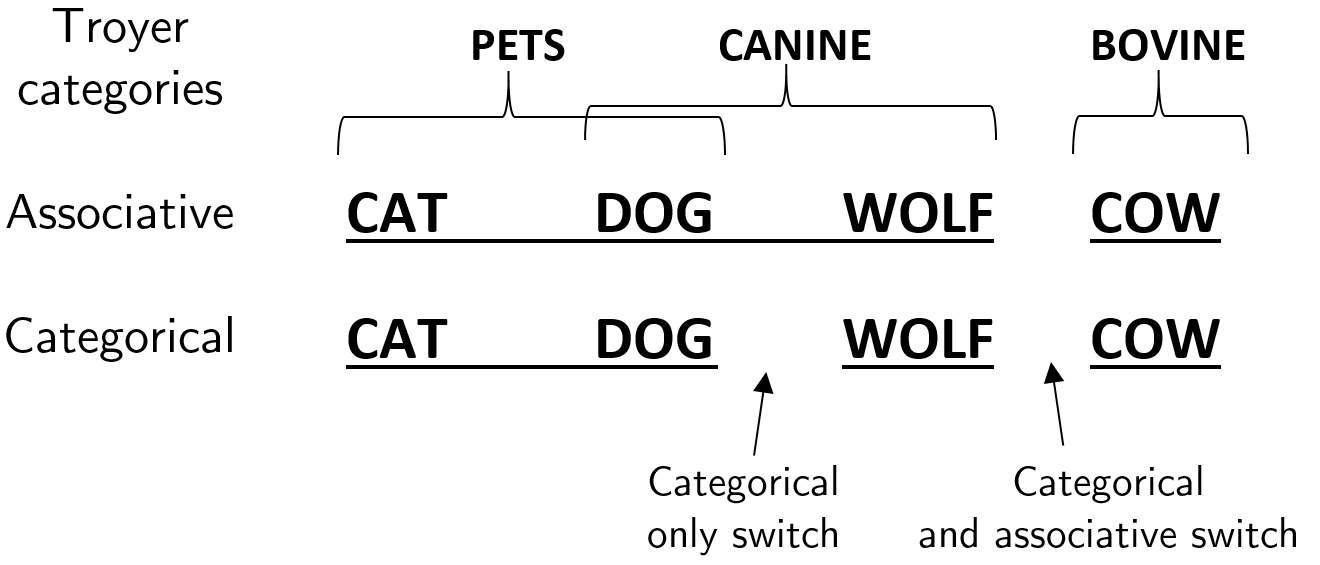}
  \vspace{-1em}
    \caption{\small{The difference between categorical and associative patch switches, based on \citeA{hills.etal.2015}.}}
    \label{fig:pst}
    \vspace{-1.5em}
\end{figure}

Each word in a random walk is labeled by the category/categories it belongs to, as defined by \citeA{troyer.etal.1997}. Words (\eg, \textsc{dog}) can belong to more than one category (\eg, \textsc{pets, canine}). As a result, there are different possibilities for defining what constitutes a patch and where the patch switches occur. We explore two different ways of defining patches over Troyer's categories, following \citeA{hills.etal.2015}, as summarized in \Figure{fig:pst}. 

\noindent\textbf{Categorical patch switch}. A patch switch occurs when a word in the sequence has no category in common with \emph{all} of the words in the current patch. In the sequence ``\textsc{cat,dog,wolf}", ``\textsc{dog,wolf}" is a patch switch because \textsc{wolf} is not in the same category as \textsc{cat} (is not a \textsc{pet}).

\noindent\textbf{Associative patch switch}. A patch switch occurs when a word in the sequence has no category in common with the \emph{last} word in the patch. For example, ``\textsc{dog,wolf}" is not a patch switch because both words share the Troyer category \textsc{canine}, but ``\textsc{wolf,cow}" is a patch switch because they have no categories in common.

From this definition it follows that all associative patch switches are also categorical patch switches. However, a categorical patch switch may not be associative; one such ``categorical only'' patch switch is illustrated in \Figure{fig:pst}.
\citeA{hills.etal.2015} argue that human search through memory is more like an associative search, and that the associative patch switch model better explains human IRT patterns. We use the associative patch switch model except where explicitly comparing the differences between the alternatives.

\section{Predicting Semantic Fluency Data}

Here we compare the results of random walks over the Batch and Incremental Networks in mimicking human semantic fluency data.  First, we focus on predicting the pattern of recall observed in human data, then we examine the properties of each patch switch model.

\subsection{Recall Patterns}
\label{sec:recallpats}

\begin{figure}
\begin{subfigure}{0.5\textwidth}
  \centering
  \includegraphics[height=4cm,width=6cm,trim=0 0 10 44, clip]{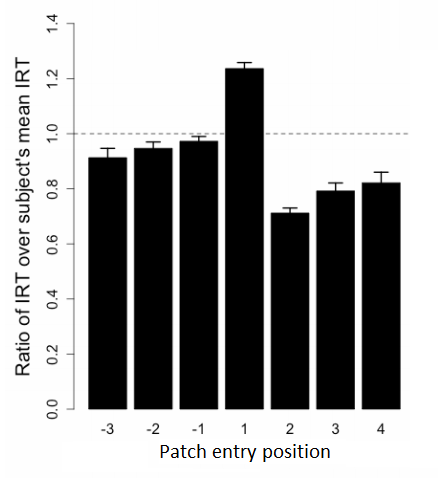}
  \caption{Human data}
  \label{fig:hills_irt}
\end{subfigure}
\qquad
\begin{subfigure}{.25\textwidth}
  \centering
  \includegraphics[trim=30 10 30 28, clip,scale=0.23]{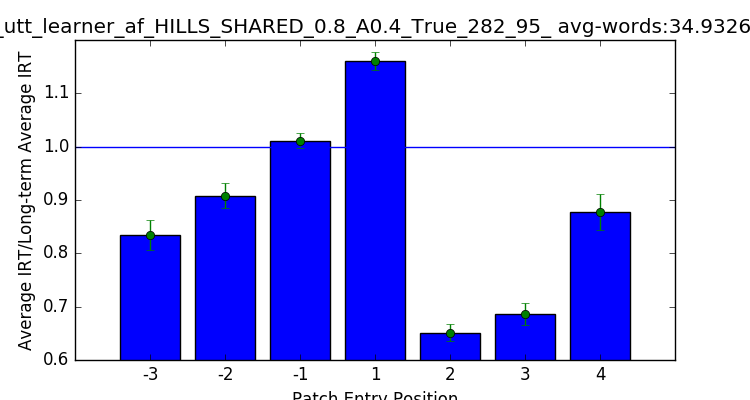}
  \caption{Batch Network}
  \label{fig:learner}
\end{subfigure}%
\begin{subfigure}{.25\textwidth}
  \centering
  \includegraphics[trim=30 10 30 28, clip,scale=0.23]{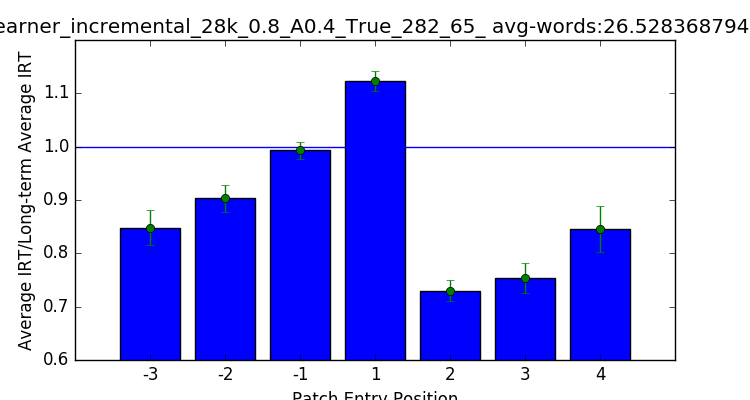}
  \caption{Incremental Network}
  \label{fig:inc-learner}
\end{subfigure}%
\caption{\small{(a) Human IRTs reproduced from \citeA{hills.etal.2012}. (b,c) IRTs from random walks generated from the simulated semantic networks. Bars are SEM.}}
\label{fig:allIRTs}
\vspace{-1.2em}
\end{figure}

\ignore{There are certain patterns present in the human semantic fluency data (\Figure{fig:hills_irt}) originally observed by \citeA{hills.etal.2012} that can be used as the basis of analysis of the goodness-of-fit for our models. In particular, }

In the human semantic fluency data (\Figure{fig:hills_irt}), the longest IRTs tend to occur between successive words that do not share a semantic category, presumably reflecting their greater distance in semantic memory \cite{hills.etal.2012,hills.etal.2015}. This is referred to as a patch switch. In the figure, a patch entry position of $1$ indicates the average IRT between the first item in a patch and the item retrieved before it. Similarly, a patch entry position of $-1$ is the average IRT between the two items preceding a patch switch. Human IRTs in patch entry position $1$ (patch switch) are higher than the average IRT, as people take longer to switch to a new patch, then dip below the average IRT at patch position $2$ as people recall words within a patch.

\fm{
As \citeA{hills.etal.2012} point out, this behavior is consistent with the marginal value theorem (MVT) of optimal foraging for patches of food in physical space \cite{charnov.1976}. In particular, MVT demonstrates that to maximize foraging gains, the optimal moment to leave a current patch is when the instantaneous reward drops below the average reward. In the human semantic search task, since participants are asked to retrieve as many words as they can, shorter IRTs lead to a bigger `reward', as more words can thus be retrieved within the time limit. Indeed, \citeA{hills.etal.2012} demonstrated that those subjects whose search patterns conformed with MVT retrieved the most words.}
\ignore{We can evaluate the average IRT for each patch entry position for our simulated walks. As in \citeA{hills.etal.2012}, we define patch switches according to the associative model.}
\fm{We evaluate whether the IRT patterns of our models also conform to the predictions of MVT as observed in the human data\ignore{, as in \citeA{nematzadeh.etal.2016}}. As such, the first patch-entry position IRT must be significantly greater than the mean IRT (e.g., the ratio between the two is greater than $1$) and all other patch entry positions must be no greater than the mean IRT. Finally, successive IRTs within the same patch should be non-decreasing.} \ignore{We consider a random walk to produce the human IRT pattern if its IRTs meet the following criteria: the ratio of the first patch-entry position IRT to the mean IRT must be at least $1.1$; for the second patch position, it must be at most $0.9$; and for all other positions, the IRT ratio must be no more than $1.0$.}

\fm{
As shown in \Figure{fig:allIRTs}, we observe a similar pattern to the human IRT data in both the Batch and Incremental Networks: the IRT drops between the first and second items in a patch, then steadily increases until the IRT exceeds the long-term average IRT, reflecting a patch switch. A single-sided t-test confirms that the first patch entry IRT is greater than the average IRT ($p\ll 0.001$). We accept the null hypothesis that the patch entry IRT at position -1 is no greater than the average IRT ($0.08\leq p\leq 0.20$). The other IRTs are significantly less than the average IRT ($p<0.02$) and successive IRTs within a patch are indeed non-decreasing.
}
\ignore{
As shown in \Figure{fig:allIRTs}, we observe a similar pattern to the human IRT data in both the Batch and Incremental Networks: the IRT drops between the first and second items in a patch, then steadily increases until the IRT exceeds the long-term average IRT, reflecting a patch switch.} %
This demonstrates, for the first time, that the combination of a simple search and structured representation that is incrementally created---simultaneously, as words are learned---can predict basic patterns observed in human semantic fluency. Next, we model additional aspects of the human data that have not been considered in previous work \cite{abbott.etal.2015, nematzadeh.etal.2016}.

\begin{figure}
\begin{subfigure}{0.5\textwidth}
  \centering
  \includegraphics[height=4cm,width=6cm,trim=0 0 10 20, clip]{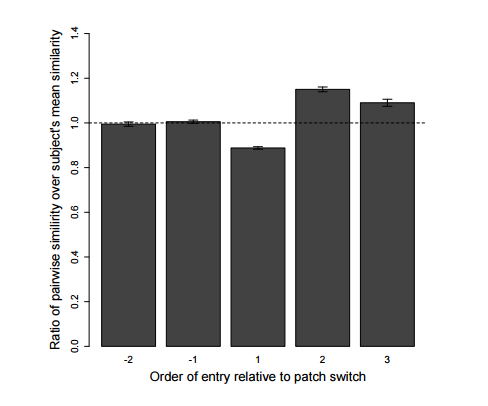}
  \caption{Human data}
  \label{fig:hills}
\end{subfigure}
\qquad
\ignore{
\begin{subfigure}{.25\textwidth}
  \centering
  \includegraphics[trim=30 10 30 28, clip,scale=0.23]{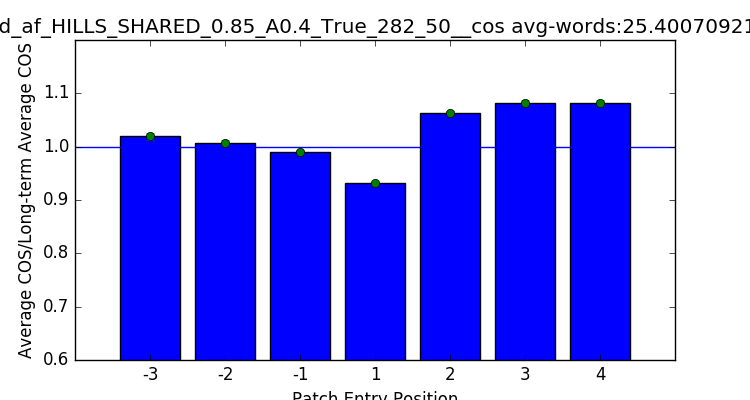}
  \caption{gold network}
  \label{fig:sfig4}
\end{subfigure}%
\begin{subfigure}{.25\textwidth}
  \centering
  \includegraphics[trim=30 10 30 28, clip,scale=0.23]{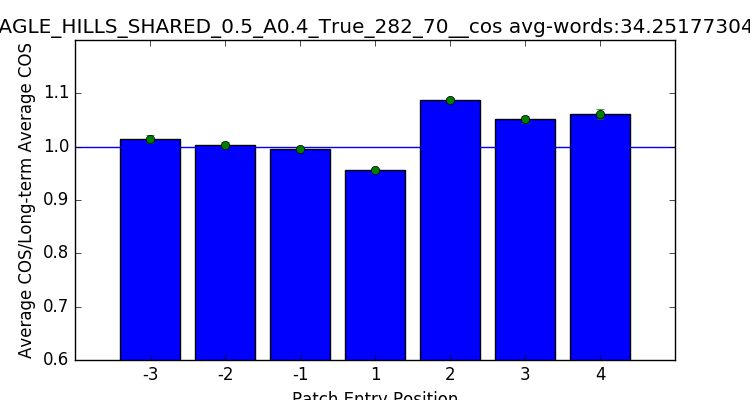}
  \caption{BEAGLE network}
  \label{fig:sfig4}
\end{subfigure}
}
\begin{subfigure}{.25\textwidth}
  \centering
  \includegraphics[trim=30 10 30 28, clip,scale=0.23]{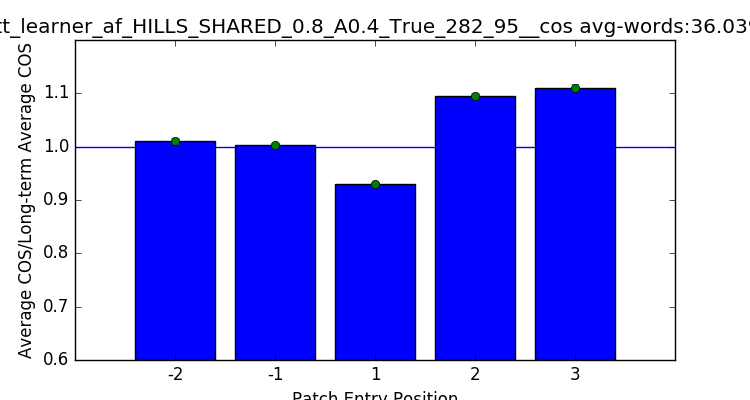}
  \caption{Batch Network}
  \label{fig:sfig3}
\end{subfigure}%
\begin{subfigure}{.25\textwidth}
  \centering
  \includegraphics[trim=30 10 30 28, clip,scale=0.23]{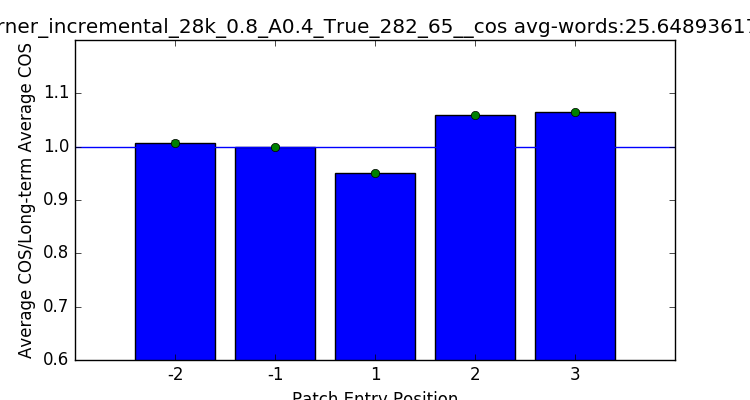}
  \caption{Incremental Network}
  \label{fig:hillsIRT}
\end{subfigure}%
\caption{\small{Cosine similarities between words in successive patch positions normalized by the average long-term cosine similarity in (a) BEAGLE vectors for items retrieved by humans \cite{hills.etal.2012}, (b,c) our semantic networks.
}}
\label{fig:allcos}
\vspace{-1.5em}
\end{figure}

A roughly analogous pattern with respect to patch entry positions is found with the average cosine similarities, although here, because cosine represents similarity rather than distance, the direction is reversed, as seen in \Figure{fig:allcos}.  Words at a patch switch are the least similar to one another. \ignore{This confirms the notion that the longer the IRT is, the less similar two words are, and the further apart they potentially are in semantic memory. Cosines are most similar at the beginning of a patch, and then similarity decreases.} \fm{Again, the first patch entry position cosine similarity is significantly less than the average cosine simimlarity ($p<0.05$). The other patch entry position cosines are on average no smaller than the average ($p\geq 0.05$). This supports the notion that words within patches are more similar (and hence, closer in semantic memory) to each other than words between patches.} \ignore{The basic qualitative features of this pattern are observed in our networks.}

\subsection{Patch Switch Type Proportion and Duration}

\begin{figure}[t]
\begin{subfigure}{0.5\textwidth}
  \centering
  \includegraphics[height=4cm,width=5.5cm,trim=0 0 0 0, clip]{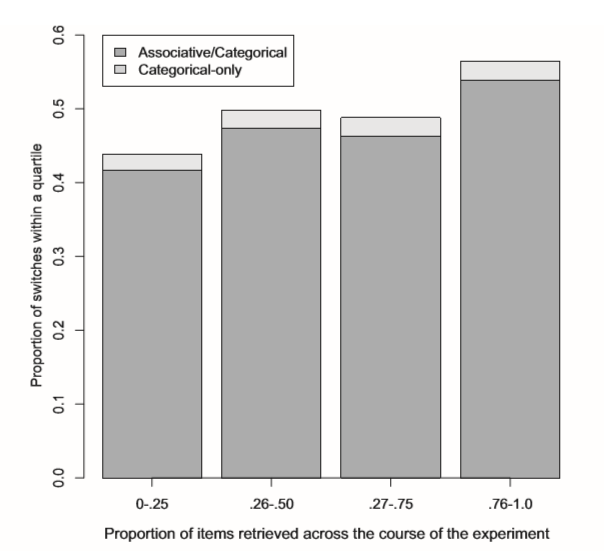}
  \caption{Human data}
  \label{fig:hills-switch}
\end{subfigure}
\qquad
\begin{subfigure}{.25\textwidth}
  \centering
  \includegraphics[trim=30 10 30 28, clip,scale=0.23]{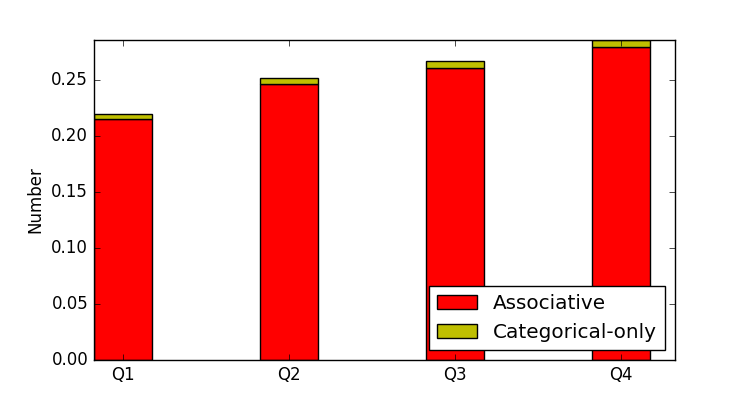}
  \caption{Batch Network}
  \label{fig:sfig3}
\end{subfigure}%
\begin{subfigure}{.25\textwidth}
  \centering
  \includegraphics[trim=30 10 30 28, clip,scale=0.23]{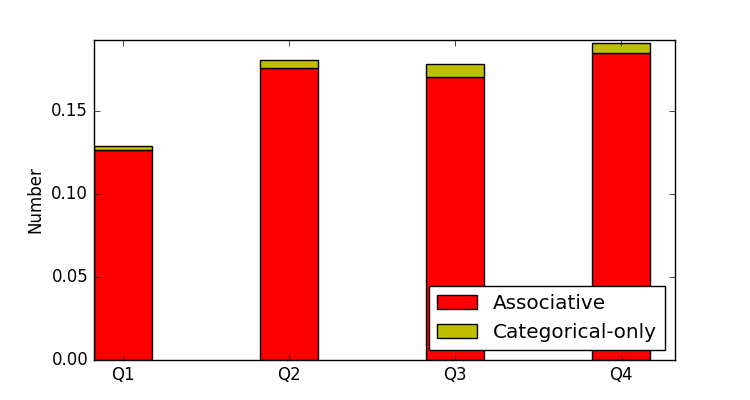}
  \caption{Incremental Network}
  \label{fig:hillsIRT}
\end{subfigure}%
\caption{\small{Average proportion of patch switch type on each quartile of the random walk for (a) human data \cite{hills.etal.2012}, (b,c) our semantic networks.}} 
\label{fig:catprop}
\vspace{-1.2em}
\end{figure}

\ignore{SS: This section needs to be simplified.  What is the human data, how do we account for it, what are one or at most two points about what's going on cognitively.  (IF the latter is relevant---does our model actually add anything to the understanding of the cognitive import of these data, or is it just that we can account for the data?)  The points about an associative vs. categorical patch switch model are not clear.}

\citeA{hills.etal.2015} categorize patch switches on the human data by whether they are associative or categorical-only (see \Figure{fig:pst}). Two observations are made from this data. 
Firstly, as in \Figure{fig:hills-switch}, the proportion of associative patch switches steadily increases throughout the four quartiles of the walk, but the number of categorical-only patch switches stays the same. 
This suggests that as more words are retrieved and semantic patches are depleted, new semantic patches must be explored. 
However, the categorical-only switches do not change in frequency. We speculate this may either be because they do not contribute to the need to explore different patches, or that they are so uncommon to begin with.

%
%
Secondly, as in Figure 5a, associative and categorical-only switches take longer than non-switches, which is expected, as non-switches search within a patch of semantically-related words. 
Associative switches take the longest, as they delineate the boundaries between the most semantically-different categories (compared to categorical-only switches).

\noindent\textbf{Model Predictions.}
When we subject the random walks on our networks to these analyses, we observe the same pattern (Figures 4,5). 
This is the first work to confirm that a random walk on semantic network is consistent with the observed pattern on the duration and proportion of different types of switches.%
%
%
%
%

\citeA{hills.etal.2015} point out the associative patch switch model has a \textit{Markov property}, insofar as that only the proceeding word's category affects the existence of a patch switch with the next word.
This is an interesting observation because it suggests that the associative switches may simply be easier to make, as only the previous word's categories affect the transition to the current word.
%
In contrast, a categorical-only switch demands higher memory overhead as the next word is affected by the overall category/categories shared by members in the current patch.
Our results show that a random walk on a structured semantic network can predict the timing and proportion of these different types of switches.

\begin{figure}
\begin{subfigure}{0.5\textwidth}
  \centering
  \includegraphics[height=4cm,width=5.5cm,trim=0 0 0 0, clip]{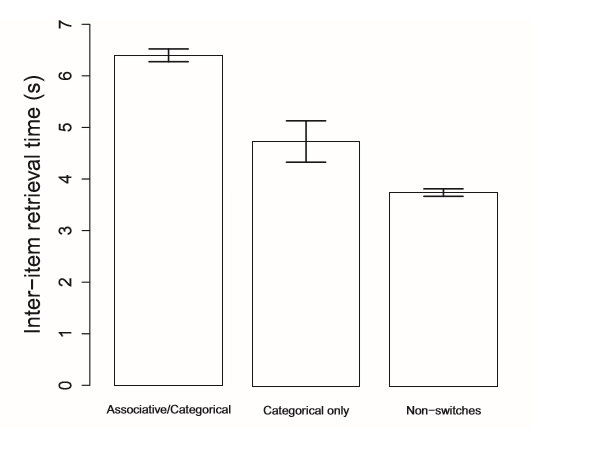}
  \caption{Human data}
  \label{fig:hills}
\end{subfigure}
\qquad
\ignore{
\begin{subfigure}{.25\textwidth}
  \centering
  \includegraphics[trim=30 10 30 28, clip,scale=0.23]{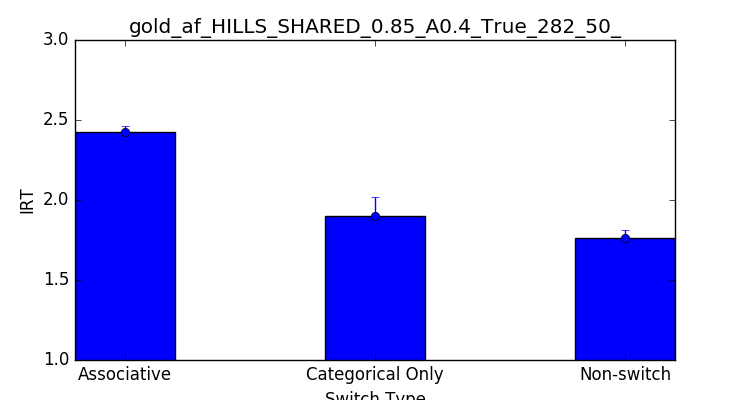}
  \caption{gold network}
  \label{fig:sfig4}
\end{subfigure}%
\begin{subfigure}{.25\textwidth}
  \centering
  \includegraphics[trim=30 10 30 28, clip,scale=0.23]{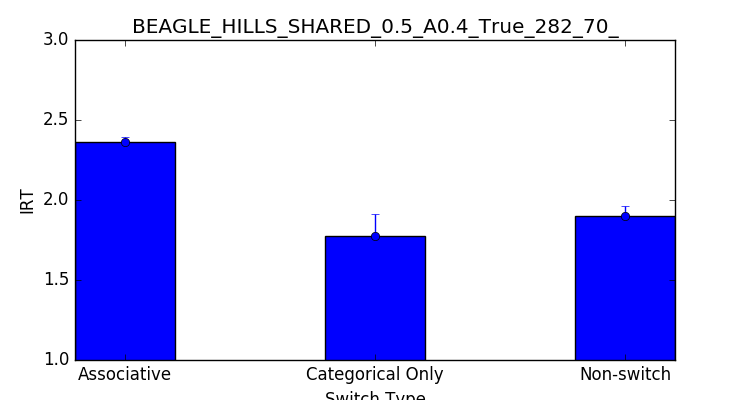}
  \caption{BEAGLE network}
  \label{fig:sfig4}
\end{subfigure}
}
\begin{subfigure}{.25\textwidth}
  \centering
  \includegraphics[trim=30 10 30 28, clip,scale=0.23]{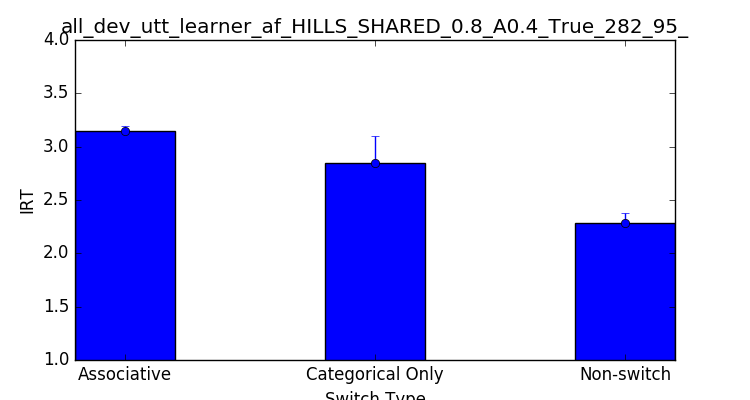}
  \caption{Batch Network}
  \label{fig:sfig3}
\end{subfigure}%
\begin{subfigure}{.25\textwidth}
  \centering
  \includegraphics[trim=30 10 30 28, clip,scale=0.23]{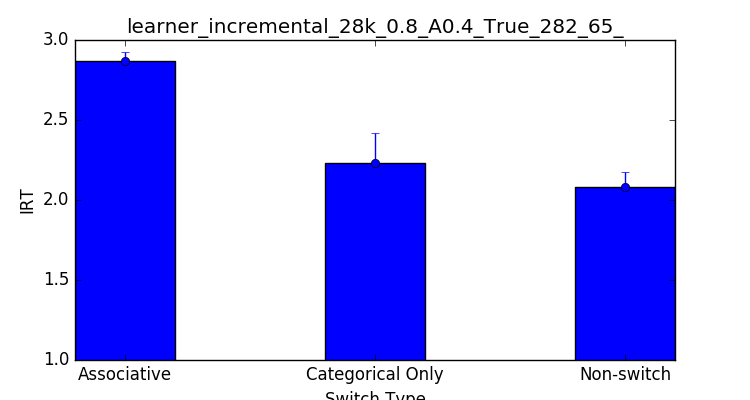}
  \caption{Incremental Network}
  \label{fig:hillsIRT}
\end{subfigure}%
\caption{\small{Average IRTs based on patch switch types for (a) human data \cite{hills.etal.2012}, (b,c) our semantic networks.}} 
\label{fig:irttype}
\vspace{-1.5em}
\end{figure}

\section{Explaining Semantic Fluency Data}
\label{sec:explain}

While our results confirm that a simple search on an incrementally-created semantic network mimics many aspects of semantic fluency behavior, not all the semantic networks predict aspects of the human data, \fm{such as adherence to MVT.}
Adding edges to the semantic network depends on the similarity between words reaching a certain threshold.
We experimented with a wide range of thresholds on similarity of word pairs (see \Section{sec:semnet-rep}) and observed \fm{that patterns consistent with MVT, as in the human IRT data (\Figure{fig:hills_irt}), appear only within a certain parameter range.} Since the choice of threshold affects the overall structure of the semantic network, we explore the features that distinguish those semantic networks that reproduce human semantic fluency patterns from those that do not.

Previous research has emphasized that semantic networks representing human knowledge have particular structural properties; namely, a small-world structure, as explained below \cite{steyvers.tenenbaum.2005}. 
However, \citeA{nematzadeh.etal.2016} observe that having a small-world structure is not a sufficient condition to guarantee a match to the observed human behavior in semantic search. 
A factor that has remained unexplored is how the quality of a network's semantic connections---whether semantically similar words are connected through a path or not---affects a network's ability to replicate findings in human semantic search. We hypothesize that this semantic quality is also important in predicting the semantic fluency data, because even two networks identical in every way except for their node labels would  produce very different behavior as the relationships between the words they represent would be completely different. 

Here we perform an extensive analysis considering both structural and semantic properties of the networks to assess which features contribute to the model's adherence to MVT, a major pattern in the human data. 
By identifying these features, we can better understand the salient aspects of semantic memory that give rise to patterns in human semantic search. We first explain how we measure the structural and semantic features of the networks. Then we discuss how we build a regression model to determine which features are responsible in predicting the semantic fluency data.

\subsection{Measuring Structure and Semantics}

A network exhibits small-world structure if it is sparse and highly connected at the same time---there are not a lot of edges in the network, but most nodes are connected through a set of high-degree nodes. As a result, the network consists of a set of highly-connected components that are connected through the high-degree nodes. Small-worldness is often quantified by $\sigma$:
\begin{align*}
    \gamma = \frac{C}{C_{random}}, \qquad \lambda = \frac{L}{L_{random}}, \qquad
    \sigma &= \frac{\gamma}{\lambda}
\end{align*}

\noindent where $C$ is the average local clustering coefficient and $L$ is the average path length, and the subscript \textit{random} refers to the metric of an equivalent Erd\H{o}s-Renyi network. A network is considered to be small-world when $\sigma>1$ (or more strictly, $\gamma\gg1$, $\lambda \approx1$) \cite{watts.strogatz.1998}. 
Intuitively, $\gamma\gg1$ reflects a structure of tightly connected components in the network, and $\lambda \approx1$ reflects relatively short path distances between nodes compared to a random network.

We observe that all of the semantic networks capable of reproducing the human patterns are small-world, but not all small-world networks generate these patterns, which is consistent with the findings of \citeA{nematzadeh.etal.2016}. As a result, we consider other structural and semantic features.  The structural features include the number of vertices ($|V|$), number of edges ($|E|$), and the sparsity of the network (average nodal degree).


\noindent\textbf{Quality of semantic connections.} 
In addition to the structure of a network, we examine the quality of its semantic connections.
We explore this by first identifying the semantic clusters formed in each network using the HDBSCAN algorithm \cite{Campello2013}, and then evaluating these clusters using Troyer's categories as our gold-standard data \citeA{troyer.etal.1997}. 
We assume that each cluster in the network can have exactly one category (e.g., pets). To determine the category label of a cluster, we examine the Troyer category memberships of each of its words, and 
assign the category label based on which category is shared by the most words of the cluster.

\ignore{
\begin{align*}
    precision&=\frac{true\text{ } positives}{true\text{ } positives+false\text{ } positives}\\ recall&=\frac{true\text{ } positives}{true\text{ } positives + false\text{ } negatives}\\
    F\text{-}score&=2\frac{precision\times recall}{precision+ recall}
\end{align*}
}
We use the standard measures of precision, recall, and F-score to assess the quality of each cluster, and average these across all clusters, weighted by cluster size, to obtain weighted precision, weighted recall, and weighted F-score for a network.  We also consider the number of clusters in each network as a feature, $|H|$.
\ignore{

In summary, we seek to characterize which of the following features most measurably contribute to the replication of patterns in human search: 

\begin{itemize}

\item \textbf{Small-world structure:} $\sigma$, $\gamma$, $\lambda$, clustering coefficient ($C$), and average path length ($L$);

\item \textbf{Overall structure:} number of vertices ($|V|$), number of edges ($|E|$), sparsity;

\item \textbf{Semantic clustering:} weighted precision, weighted recall, weighted F-score, number of clusters determined by HDBSCAN ($|H|$).
\end{itemize}
}

\subsection{Analyzing the Contribution of Features}

We characterize which structural and semantic features of a network are most important (in predicting human data) by fitting logistic regression models on all the possible combinations of features.

Prior to training, feature values were transformed into z-scores (i.e., for a given feature $x$ for a given network $i$, the standardized value is $(x_i-\bar{x}) / \hat{s}$; $\bar{x}$ is the sample mean of the feature for all networks and $\hat{s}$ is sample standard deviation). This permits the coefficients of regression to be compared directly in terms of their contribution in predicting the data.\footnote{Although some of these features are dependent (\eg, $|E|$ and $sparsity$), we do not include their interactions in our regression analysis. We focus on understanding whether a subset of individual features can explain the human data and thus examine all possible combinations of features.}

\subsubsection{Experimental Set-Up}

Logistic classifier models were trained on a set of Batch and Incremental networks. \fm{During training, we ensure an equal representation of networks that adhere to and do not adhere to MVT. This is a binary condition satisfied according to the criteria explained in \Section{sec:recallpats}.} Networks were first generated across the entire parameter space of the similarity thresholds (i.e., all combinations of $\rho$ and $\rho_{animal}$ ranging from 0 to 1, in increments of 0.1). We excluded networks where the number of nodes reachable by the starting word `animal' was smaller than 30, as they would not be able to produce as many words as human participants did ($37\pm5$) \cite{hills.etal.2012}. Since the number of non-IRT producing networks outnumbered the IRT producing networks, we uniformly sampled the parameter space in which IRT pattern-producing networks occurred so that the number of each would be equal.
Using this procedure, 42 Batch and 56 Incremental networks were generated. In each case, exactly half of the networks produce the IRT pattern \fm{consistent with MVT}. 

\noindent\textbf{Model selection.} For each set of Batch and Incremental networks, we examine which features best predict the human data by building and evaluating logistic regression models for all possible combinations of the features. Model selection was performed in two steps. First, the models with the highest stratified-3-fold (SKF) cross-validation score were taken. From these, the model with the fewest number of features was selected. 

\subsubsection{Results of Logistic Regression}

\Table{tab:lr} shows the features that appeared in the logistic regression model that achieved the best SKF cross-validation score for each of the types of networks. Since each feature was standardized (with $mean=0$ and $variance=1$), the magnitude of the coefficients can be interpreted directly. We note that small-worldness ($\sigma$) and weighted F-score are influential predictors for both Batch and Incremental networks. In both models, weighted F-score is the most influential predictor. Although $\sigma$ is the least influential predictor, we find it significant that it is a shared predictor for both networks. Structural properties relating to the number of edges ($|E|,sparsity$) as well as clustering coefficient ($C,\gamma$), are structural properties that have been previously characterized in semantic networks \cite{steyvers.tenenbaum.2005,goni.etal.2010}.
\ignore{
The original criteria for small-worldness were defined as $\lambda\approx 1,\gamma\gg1$ \cite{watts.strogatz.1998}, which respectively refer to a low average path length, but high degree of clustering compared to a random network. Across all our networks, the $\gamma$ term varies more than $\lambda$ as $\lambda\approx 1$. \sxs{So why is lambda one of the best predictors for Batch?} The result of training the logistic classifier suggests that $\gamma$ is predictive of the IRT pattern for all networks. Since $\gamma$ is the dominating term in $\sigma$, this observation is consistent with previous findings regarding small-world structure in semantic networks \cite{steyvers.tenenbaum.2005,nematzadeh.etal.2016}.
In addition to the small-worldness criterion, we observe that the weighted F-score is a shared predictor.} Hence, we conclude that both topological features---namely, small worldness (high clustering coefficient and short average path length)---and semantic features---high weighted F-score (good precision and recall in clusters)---are jointly associated with reproducing the IRT pattern. 

\ignore{Not sure if the above is old text or I'm just not understanding it.  It seems like for both Batch and Incremental, small-worldness and some of its components are important; plus sparsity of the graph (if number of edges and sparsity are inversely affecting the regression; otherwise I don't understand this, since greater sparsity means fewer edges); and weighted F-score.}


\begin{table}
\begin{center}
\resizebox{\columnwidth}{!}{%
\begin{tabular}{ |l| r |*5c|}  
\ignore{
\multicolumn{1}{c}{}  & \multicolumn{3}{|c|}{Structure} &\multicolumn{3}{|c|}{Semantics} \\ }
\hline
  \multicolumn{1}{|c|}{Networks} &  \multicolumn{1}{c}{Acc.} & \multicolumn{5}{|c|}{Features and Coefficients} \\
\hhline{|=|=|=====|}
 Batch       & 93\% & \pmb{$\sigma$}& $\lambda$& C& sparsity& \textbf{weighted F-score}
 \\
 &&\textbf{0.58}&0.74&-1.92&0.94&\textbf{0.94}\\
 \hline
 Incremental &  90\% & &\pmb{$\sigma$}& $\gamma$& $|E|$& \textbf{weighted F-score} \\
 &&&\textbf{0.65}&0.71&-1.64&\textbf{1.07}\\
\hline
\end{tabular}
}
\end{center}
\vspace{-0.3cm}
\caption{\small{Features used to train the logistic regression models for predicting IRT pattern production with the highest stratified 3-fold cross-validation accuracy (Acc.). Shared features are bolded.}}
\label{tab:lr}
\vspace{-0.5cm}
\end{table}
\section{Conclusions}

Learning word meanings and representing them in semantic memory are processes that often occur simultaneously, notably in early language acquisition.  A cognitive model capable of integrating these two processes will therefore more realistically capture language acquisition and usage. \fm{It is noteworthy that both the Batch and Incremental Networks perform comparably on all of the data examined here. We consider this strong support for the hypothesis that semantic networks learned incrementally on a naturalistic language corpus can replicate search patterns in the free recall task, a claim that is neither obvious nor trivial to demonstrate. Furthermore, some of the performance characteristics we use in measuring the fit of the model to the human data---namely, whether the IRT patterns produced by the model are consistent with MVT or not---are binary conditions: either the behavior is replicated or it is not, so, barring additional criteria, a graded scale by which to score performance is not possible. Future work will seek to better characterize the performance differences between the two models.}

We deploy a model that can generate semantic networks incrementally from naturalistic language use, \ie \ child-directed speech, while it gradually learns the word meanings, lending it plausibility as a cognitive model. We show this model replicates human performance on semantic fluency tasks; namely, with regards to patch entry IRT, patch entry cosine similarity patterns, patch switch type proportions, and patch switch type IRTs. We show, furthermore, that the Markov property of the random walk does indeed align with the associative nature of search in the human semantic fluency task \cite{hills.etal.2015}.

By investigating the structural and semantic features of these and other networks, we show that small-worldness alone does not explain the ability of a network to replicate the human patterns. Having highly connected components, \textit{and} ones that reflect the semantic categories of words, are both properties that may be necessary in
predicting semantic search behavior observed in humans.


\ignore{
One further question of interest is whether patch switches in semantic memory are captured adequately by the fluid patch model, or whether an alternative model may be provide a better fit to human data (e.g., a static patch model, in which a patch switch occurs whenever a word shares no category in common with \textit{all} words in the current patch rather than simply the last one) \cite{hills.etal.2015}. We seek to explore this question in future work.
}


\bibliographystyle{acl_natbib}
\bibliography{nematzadeh,sw}

\end{document}